# Dual Memory Neural Computer for Asynchronous Two-view Sequential Learning

Hung Le, Truyen Tran and Svetha Venkatesh
Applied AI Institute, Deakin University, Geelong, Australia
{lethai,truyen.tran,svetha.venkatesh}@deakin.edu.au

February 11, 2018

**Abstract**

One of the core tasks in multi-view learning is to capture relations among views. For sequential data, the relations not only span across views, but also extend throughout the view length to form long-term intra-view and inter-view interactions. In this paper, we present a new memory augmented neural network model that aims to model these complex interactions between two asynchronous sequential views. Our model uses two encoders for reading from and writing to two external memories for encoding input views. The intra-view interactions and the long-term dependencies are captured by the use of memories during this encoding process. There are two modes of memory accessing in our system: late-fusion and early-fusion, corresponding to late and early inter-view interactions. In the late-fusion mode, the two memories are separated, containing only view-specific contents. In the early-fusion mode, the two memories share the same addressing space, allowing cross-memory accessing. In both cases, the knowledge from the memories will be combined by a decoder to make predictions over the output space. The resulting dual memory neural computer is demonstrated on a comprehensive set of experiments, including a synthetic task of summing two sequences and the tasks of drug prescription and disease progression in healthcare. The results demonstrate competitive performance over both traditional algorithms and deep learning methods designed for multi-view problems.

## 1 Introduction

In multi-view learning, data can be naturally partitioned into channels presenting different views of the same data. For examples, multilingual documents have one view for each language, and images of a 3D object are taken from different viewpoints. Multi-view sequential learning is a sub-class of multi-view learning where each view data is in the form of sequential events, which can be synchronous or asynchronous. In the synchronous setting, all views share the same time step. Some problems of this type include video consisting of visual and audio streams; and text as a joint sequence of words and part-of-speech tags. Synchronous multi-view sequential learning is an active area [20, 26, 27]. Despite their practical usages, these works make assumptions on the time step alignment and thus they are constrained by the scope of synchronous multi-view problems.

In this work, we relax these assumptions and focus more on asynchronous sequential multi-view, that is, there is no alignment among views and the sequence lengths vary across views. These occur when the data is collected from channels having different time scales or we cannot infer the precise time information when extracting data. In healthcare, for instance, an electronic medical record (EMR) contains information on patient's admissions, each of which consists of various views such as diagnosis, medical procedure, and medicine. Although an admission is time-stamped, medical events from each view inside the admission are not synchronous and different in length.

Asynchronous multi-view data often demonstrates three types of view interactions. The first type is intra-view interactions, those involving only one view, representing the internal dynamics. For examples, each EMR view has specific rules for coding its events, forming distinctive correlations among medical events inside a particular view. The second type is late inter-view interactions, those that span from input views to output, representing the mapping function between the inputs and the outputs. We call it "late" because the interaction across input views is considered only in the inference process. The third type is early inter-view interactions, those that account for relations covering multiple input views and happening before the inference process. For example, in drug prescription, the diagnosis view is the cause of the medical procedure view, both of which affect the output which are medicines prescribed for patient. The interactions in sequential views not only span across views but also extend throughout the length of the sequences. One example involves patients whose diseases in current admission are related to other diseases or treatments from distant admissions in the past. The complexity of view interactions, together with the unalignment and long-term dependencies among views poses a great challenge in asynchronous sequential multi-view problems.

We propose a novel memory augmented neural network model (MANN) solving the problem of asynchronous interactions and long-term dependencies at the same time. Our model makes use of three neural controllers and two external memories constituting a dual memory neural computer. In our architecture, each input view is assigned to a controller and a memory to model the intra-view interactions in that particular view. At each time step, the controller reads an input event, updates the memory, and generates an output based on its current hidden state and read vectors from the memory. Corresponding to the two types of inter-view interactions, there are two modes in our architecture: late-fusion and early-fusion memories. In the late-fusion mode, the memory space for each view is separated and independent, that is, there is no information exchange between the two memories during the encoding process. The memories' read values are only synthesized to generate inter-view knowledge in the decoding phase. Contrast to the late-fusion mode, the memory addressing space in the early-fusion mode is shared among views. That is, the encoder from one view can access and modify the contents of the other view's memory. This design ensures the information is shared across views via memories accessing. In order to facilitate this asynchronous sharing, we design novel cache components that temporarily hold the write values of every time steps. This enables related information at different time steps to be written to the memories together. Finally, we apply memory write-protected mechanism in the decoding process to make the inference more efficient.

In summary, our main contributions are: (i) proposing a novel dual memory neural computer (DMNC) to solve the asynchronous multi-view sequential problem, (ii) designing our architecture to model view interactions and long-term dependencies, (iii) demonstrating the efficacy of our proposed model on real-world medical data sets for the problems of drug prescription and disease progression. The significance of DMNC lies in its versatility as our model presents a generic approach that uses external memories to model multi-view problems.

## 2 Background

### 2.1 Related Works

Multi-view learning is a well-studied problem, where methods often exploit either the consensus or the complementary principle [25]. A straightforward approach is to concatenate all multiple views into one single view making it suitable for conventional machine learning algorithms, both for vector inputs [6, 28] or sequential inputs [12, 23]. Another approach is co-training [2, 14], aiming to maximize the mutual agreement on views. Other approaches either establish a latent subspace shared by multiple views [19] or perform multiple kernel learning [21]. These works are typically limited to non-sequential views.

More recently, deep learning is increasingly applied for multi-view problems, especially with sequential data. For examples, LSTM [9] is extended for multi-view problems in [20] or multiple kernel learning is combined with convolution networks in [16]. More recent methods focus on building deep networks to extract features from each view before applying different late-fusion techniques such as tensor products [27], contextual LSTM [17] and gated memory [26]. All of these deep learning methods are designed only for synchronous sequential input views. Hence, the applications of these methods mostly fall into tagging problems where the output is aligned with the input views. As far as we know, the only work that can apply to asynchronous inputs is [5], in which the authors construct a dual LSTM for feature extraction and use attention for late-fusion.

In healthcare, there are only few works that make use of multi-view data. A multi-view multi-task model is proposed in [13] to predict future chronic diseases given multi-media and multi-model observations. However, this model is only designed for single-instance regression problems. DeepCare [15] solves the disease progression problem by combining diagnosis and intervention views. It treats events in each admission as a bag and use poolings to compute the feature vectors for the two views in an admission. The sequential property of events inside each admission is ignored and there is no mechanism to model inter-view interactions at event level. There are many other works using deep learning such as RETAIN [4], Dipole [11] and LEAP [29] that attack different problems in healthcare. However, they are designed for single input view.

Memory augmented neural network (MANN) is a recent promising research topic in deep learning. Memory Networks (MemNNs) [24] and Neural Turing Machines (NTMs) [7] are the two classes of MANNs which have been applied to many problems including healthcare [18]. However, designing a MANN for multi-view learning is still new and our work is one of the first attempts to build a generic MANN capable of modeling interactions among events from different data views. The memories used in our model are based on the powerful DNC [8], the latest improvement over the NTM. Since DNC is the building block in our model, we briefly present it in the next subsection.

## 2.2 DNC Overview

A DNC consists of a controller, which accesses and modifies an external memory module using a number of read and write heads. Given some input $x_t$, and a set of $R$ previous read values from memory $r_{t-1} = \left[r_{t-1}^1, r_{t-1}^2, ..., r_{t-1}^R\right]$, the controller produces a key $k_t \in \mathbb{R}^D$, where $D$ is the word size in memory. This key will be used to compute content-based read-weight and write-weight vector for a memory matrix $M_t \in \mathbb{R}^{N \times D}$, where $N$ is the number of memory locations. In addition to content-based addressing, DNC supports dynamic memory allocation and temporal memory linkage for computing the final write-weight $w_t^w$ and read-weights $w_t^{rk}$. The memory is updated by following rule:

$$M_t = M_{t-1} \circ \left(E - g_t^w w_t^w e_t^\top\right) + g_t^w w_t^w v_t^\top \qquad (1)$$

where $E$ is an $N \times D$ matrix of ones, $g_t^w$ is a scalar write gate, $w_t^w \in [0,1]^N$ is the final write-weight, $e_t \in [0,1]^D$ is an erase vector, $v_t \in \mathbb{R}^D$ is a write vector, $\circ$ is point-wise multiplication. The $k$-th read value $r_t^k$ is retrieved using:

$$r_t^k = M_t^\top w_t^{rk}, 1 \leq k \leq R \qquad (2)$$

## 3 Methods

### 3.1 Asynchronous Two-View Sequential Learning: Problem Formulation

Let us start with a generic formulation of the asynchronous two-view sequential learning problem. Let $S^{i_1}$, $S^{i_2}$ denote the two input view spaces and $S$ denotes the output view space. Each sample of the two-view problem $\left(X^{i_1}, X^{i_1}, Y\right)$ consist of two input views: $X^{i_1} = \{x_1^{i_1}, ..., x_{t_1}^{i_1}, ..., x_{L^{i_i}}^{i_1}\}$, $X^{i_2} = \{x_1^{i_2}, ..., x_{t_2}^{i_2}, ..., x_{L^{i_2}}^{i_2}\}$ and one output view $Y = \{y_1, ..., y_t, ..., y_L\}$ – where each view can have different lengths ($L^{i_1}$, $L^{i_2}$, $L$) and can be seen as a set/sequence of events that belongs to different spaces ($x_{t_1}^{i_1} \in S^{i_1}$, $x_{t_2}^{i_2} \in S^{i_2}$, $y_t \in S$). Each event then can be represented by an one-hot vector $v \in [0, 1]^{\|C\|}$, where $C$ can be $S^{i_1}$, $S^{i_2}$ or $S$. It should be noted that this formulation can be applied to many situations including video-audio understanding, image-captioning and other two-channel time-series signals. Here we focus effort on solving the two-view problems in healthcare.

For our healthcare problems, we restrict the scope to modeling Electronic Medical Record (EMR), which typically contains the history of hospital encounters, including diagnoses and interventions such as procedures and drugs. In drug prescription, doctors prescribe drugs after considering diagnoses and procedures administered to patients. In modeling disease progression, doctor may refer to patient's history of admissions to help diagnoses the current disease or to predict the future disease occurrences of the patient. There are clinical recording rules applying to EMR codes such that diagnoses are "ordered by priority" or procedures follow the order that "the procedures were performed"[1]. Besides, although medical codes from different views are highly correlated, they are not aligned. For instances, some diagnoses may correspond to one procedure or one diagnosis may result in multiple medicines. Hence, these problems can be treated as asynchronous two-view sequential learning.

In the drug prescription context, $S^{i_1}$ and $S^{i_2}$ represent the diagnosis and procedure spaces, respectively and $S$ corresponds to the medicine space. The drug prescription objective is to select an optimal subset of medications from $S$ based on diagnosis and procedure codes. Similarly, we can formulate the disease progression problem as two input sequences (diagnoses and interventions) and one output set (next diagnoses). Although our architecture can model sequential output, the choice of representing output as set is to follow a common practice in healthcare where the order of medical suggestions is specified. Because a patient may have multiple admission records for different hospital visits, a patient record can be represented as $\left\{ \left( X_a^{i_1}, X_a^{i_1}, Y_a \right) \right\}_{a=1}^{A}$, where $A$ is the number of admissions this patient commits. In order to predict $Y_a$, we may need to exploit not only $\left( X_a^{i_1}, X_a^{i_1} \right)$ but also $\left\{ \left( X_{pa}^{i_1}, X_{pa}^{i_1} \right) \right\}_{pa=1}^{a-1}$. More details on how our work makes use of previous admissions and handle long-term dependencies will be given in Section 3.4.

## 3.2 Dual Memory Neural Computer

We now present our main contribution to solve the generic asynchronous two-view sequential learning – a new deep memory augmented neural network called Dual Memory Neural Computer (DMNC) (see Fig. 1). Our architecture consists of three neural controllers (two for encoding and one for decoding), each of which interacts with two external memory modules. Each of the two memory modules is similar to the external memory module in DNC [8], that is they are equipped with temporal linkage and dynamic allocation. The three controllers have their own embedding matrices $W_E^{i_1}$, $W_E^{i_2}$, $W_E$ which project the one-hot representation of events to a unified $d$-dimensional space. We use $\mathbf{x}_{t_1}^{i_1}$, $\mathbf{x}_{t_2}^{i_2}$, $\mathbf{y}_t$ $\in \mathbb{R}^d$ to denote the embedding vector of $x_{t_1}^{i_1}$, $x_{t_2}^{i_2}$, $y_t$, respectively, in which $\mathbf{x}_{t_1}^{i_1} = W_E^{i_1} x_{t_1}^{i_1}, \mathbf{x}_{t_2}^{i_2} = W_E^{i_2} x_{t_2}^{i_2}, \mathbf{y}_t = W_E y_t$. The embedding vectors $\mathbf{x}_{t_1}^{i_1}$, $\mathbf{x}_{t_2}^{i_2}$ are always used as inputs of the encoders while the embedding vector $\mathbf{y}_t$ will only be used as input of the decoder if the output view is a sequence.

---
[1] https://mimic.physionet.org/mimictables/

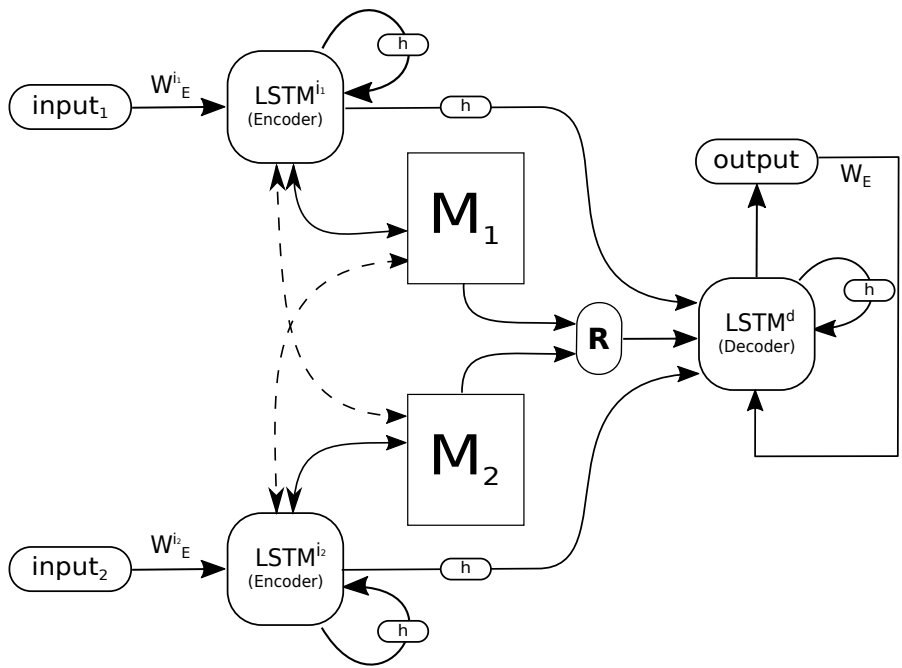

Figure 1: Dual Memory Neural Computer. $LSTM^{i_1}$, $LSTM^{i_2}$ are the two encoding controllers implemented as LSTMs. $LSTM^d$ is the decoding controller. The dash arrows represent cross-memory accessing in early-fusion mode.

Each encoder will transform the embedding vectors to $h$-dimensional hidden vectors. The current hidden vectors and outputs of the encoders are computed as:

$$h_{t_1}^{i_1}, o_{t_1}^{i_1} = LSTM^{i_1}\left(\left[\mathbf{x}_{t_1}^{i_1}, r_{t_1-1}^{i_1}\right], h_{t_1-1}^{i_1}\right), 1 \leq t_1 < L^{i_1} \tag{3}$$

$$h_{t_2}^{i_2}, o_{t_2}^{i_2} = LSTM^{i_2}\left(\left[\mathbf{x}_{t_2}^{i_2}, r_{t_2-1}^{i_2}\right], h_{t_2-1}^{i_2}\right), 1 \leq t_2 < L^{i_2} \tag{4}$$

where $r_{t_1-1}^{i_1}$, $r_{t_2-1}^{i_2}$ are read vectors at previous time step of each encoder and $L^{i_1}, L^{i_2}$ are the lengths of input views. It should be noted that the time step in each view may be asynchronous and the lengths may be different. In our applications, since we treat input views as sequences, we use $LSTM$ as the core of the encoders[2]. Using separated encoder for each view naturally encourages the intra-view interactions. To model inter-view interactions, we use two modes of memories, late-fusion and early-fusion.

**Late-fusion memories:** In this mode, our architecture only models late inter-view interactions. In particular, $r_{t_1}^{i_1}$ and $r_{t_2}^{i_2}$ are computed separately:

$$r_{t_1}^{i_1} = \left[r_{t_1}^{i_1,1}, ..., r_{t_1}^{i_1,R}\right] = m_{read}^{e_1}\left(o_{t_1}^{i_1}, M_1\right) \tag{5}$$

$$r_{t_2}^{i_2} = \left[r_{t_2}^{i_2,1}, ..., r_{t_2}^{i_2,R}\right] = m_{read}^{e_2}\left(o_{t_2}^{i_2}, M_2\right) \tag{6}$$

where $M_1$, $M_2$ are the two memory matrices containing view-specific contents and $m_{read}^{e_1}$, $m_{read}^{e_2}$ are two read functions of the encoders with separated set of parameters. Given the controller output vectors, the read functions produce the keys $k_{t_1}^{i_1}$, $k_{t_2}^{i_2}$ in the manner of DNC. The keys are used to address the corresponding memory and compute the read vectors using Eq.(2). This design ensures the dynamics of computation in one view does not affect the other's and only in-view contents are stored in view-specific memory. This mode is important because in certain situations, writing external contents to view-specific memory will interfere the acquired knowledge and obstruct the learning process. In Section 4.1, we will show a case study that fits with this setting and the empirical results will demonstrate that the late-fusion mode is necessary to achieve better performance.

**Early-fusion memories:** When there exists a strong correlation between the two input views, requiring to model early inter-view interactions, we introduce another mode of memories: early-fusion mode. In this mode, the two memories share the same addressing space, that is, the encoder from one view can access the memory content from another view and vice versa. Also, the read functions $m_{read}^e$ share the same parameter set:

$$r_{t_1}^{i_1} = \left[r_{t_1}^{i_1,1}, ..., r_{t_1}^{i_1,R}\right] = m_{read}^e\left(o_{t_1}^{i_1}, [M_1, M_2]\right) \quad (7)$$

$$r_{t_2}^{i_2} = \left[r_{t_2}^{i_2,1}, ..., r_{t_2}^{i_2,R}\right] = m_{read}^e\left(o_{t_2}^{i_2}, [M_1, M_2]\right) \quad (8)$$

---

[2] For inputs as sets, we can replace the $LSTMs$ with $MLPs$

**Algorithm 1** Training algorithm for healthcare data (set output)
**Require:** Training set $\{\{(X_a^{i_1}, X_a^{i_2}, Y_a\}_{a=1}^{A}\}_{n=1}^{N}$
1: Sample $B$ samples from training set
2: **for each** sample in $B$ **do**
3:     Clear memory $M_1, M_2$
4:     **for** $a = 1, A$ **do**
5:         $(X^{i_1}, X^{i_2}, Y) = (X_a^{i_1}, X_a^{i_2}, Y_a)$
6:         **while** $t_1 < L^{i_1}$ or $t_2 < L^{i_2}$ **do**
7:             **if** $t_1 < L^{i_1}$ **then**
8:                 Use $Eq.(3)$ to calculate $h_{t_1}^{i_1}, o_{t_1}^{i_1}$
9:                 Use $Eq.(1)$ or $Eq.(11)$ to update $M_1$
10:               Use $Eq.(5)$ or $Eq.(7)$ to read $M_1$
11:               $t_1 = t_1 + 1$
12:             **end if**
13:             **if** $t_2 < L^{i_2}$ **then**
14:               Use $Eq.(4)$ to calculate $h_{t_2}^{i_2}, o_{t_2}^{i_2}$
15:               Use $Eq.(1)$ or $Eq.(11)$ to update $M_2$
16:               Use $Eq.(6)$ or $Eq.(8)$ to read $M_2$
17:               $t_2 = t_2 + 1$
18:             **end if**
19:         **end while**
20:         Use $Eq.(12)$ and $Eq.(13)$ to read $M_1, M_2$
21:         Use $Eq.(16)$ to calculate $\widehat{y}$
22:         Update parameter $\theta$ using $\nabla_\theta Loss_{set}(Y, \widehat{y})$
23:     **end for**
24: **end for**

Since the read vectors for one encoder can come from either memories, the encoder's next hidden values are dependent on both views' memory contents, which enable possible early inter-view interactions in this mode.

**Memories modification with cache components:** In both modes, the two memories are updated every time step by the two encoders. While in the late-fusion mode, the writing to two memories are independent and can be executed in parallel using Eq.(1), in the early-fusion mode, the writings must be executed in an alternating manner. In particular, the two encoders take turn writing to memories, allowing the exchange of information at every time step. Doing this way is optimal if the two views are synchronous and equal in lengths. To make it work with variable length input views, we introduce
a new component to our architecture: a cache memory that lies between the
controller and the external memory. Different from the original DNC which
writes directly the event's value to the external memory, in the early-fusion
mode of our architecture, each controller integrates write values inside its own
cache memory $c_t$ until an appropriate moment before committing them to the
external memory. We introduce $g_t^c$ as a learnable cache gate to control the degree
of integration between current write value and the previous cache's content as
follows:

$$g_t^c = f^c\left(o_t^i\right) \tag{9}$$
$$c_t = g_t^c \circ c_{t-1} + (1 - g_t^c) \circ v_t \tag{10}$$

In these equations, $g_t^c$ is the cache gate, $o_t^i$ is the encoder output, $f^c$ is a
learnable function[3], $c_t$ is the cache content and $v_t$ is the write value. Then, the
cache will be written to the memory using the following formula:

$$M_t = M_{t-1} \circ \left(E - g_t^w w_t^w e_t^\top\right) + g_t^w w_t^w c_t^\top \tag{11}$$

We propose this new writing mechanism for early-fusion mode to enable
one encoder to wait for another while processing input events (in this context,
waiting means the encoder stops writing to memory). In the original DNC, if
the write gate $g_t^w$ is close to zero, the encoder does not write to memory and
the write value at current time step will be lost. However, in our design, even
when there is no writing, the write value somehow can be kept in the cache if
$g_t^c < 1$. The cache in a view may choose to hold an event's write value instead
of writing it immediately at the read time step. Thus, the information of the
event is compressed in the cache until appropriate occasion, which may be after
the appearance of another event from the other view. This mechanism enables
two related asynchronous events to simultaneously involve in building up the
memories.

**Write-protected memories:** In our architecture, during the inference
process, the decoder stops writing to memories. We add this feature to our
design because the decoder does not receive any new input when producing
output. Writing to memories in this phase may deteriorate the memory contents,
hampering the efficiency of the model.

### 3.3 Inference in DMNC

In this section, we give more details on the operation of the decoder. Because
the decoder works differently for different output types (set or sequence), we
will present two versions of decoder implementation.

**Output as sequence:** In this setting, the decoder ingests the encoders' final
states as its initial hidden state $h_0 = \left[h_{L^{i_1}}^{i_1}, h_{L^{i_2}}^{i_2}\right]$ . The decoder's hidden and
output vectors are given as: $h_t, \left[o_t^1, o_t^2\right] = LSTM^d\left(\left[\mathbf{y}_{t-1}^*, r_{t-1}^{i_1}, r_{t-1}^{i_2}\right], h_{t-1}\right)$.

---
[3]In this paper, all $f$ functions are implemented as single-layer feed-forward neural networks

Here, $\mathbf{y}^*_{t-1}$ is the embedding of the previous prediction $y^*_{t-1}$. The decoder combines the read vectors from both memories to produce a probability distribution over the output:

$$r_t^{i_1} = \left[r_t^{i_1,1}, ..., r_t^{i_1,R}\right] = m_{read}^d\left(o_t^1, M_1\right) \tag{12}$$

$$r_t^{i_2} = \left[r_t^{i_2,1}, ..., r_t^{i_2,R}\right] = m_{read}^d\left(o_t^2, M_2\right) \tag{13}$$

$$P\left(y_t|X^{i_1}, X^{i_2}\right) = \pi\left(\left[o_t^1, o_t^2\right] + f^d\left(\left[r_t^{i_1}, r_t^{i_2}\right]\right)\right) \tag{14}$$

where $r_t^{i_1}, r_t^{i_2}$ are read vector from $M_1, M_2$, respectively, provided by the read function $m_{read}^d$, $f^d$ is a learnable function and $\pi$ is softmax function. Then, the current prediction is $y_t^* = \underset{y \in S}{argmax}\, P\left(y_t = y|X^{i_1}, X^{i_2}\right)$ and the loss function is the cross entropy:

$$Loss_{seq}\left(Y, P\right) = -\sum_{t=1}^{L} \log P\left(y_t|X^{i_1}, X^{i_2}\right) \tag{15}$$

**Output as set:** In this setting, the decoder uses $m_{read}^d$ to read from the memories once to get the read vectors $r^{i_1}, r^{i_2}$. The decoder combines these vectors with the encoders' final hidden values to produce the output vector $\widehat{y} \in \mathbb{R}^{|S|}$:

$$\widehat{y} = \sigma\left(f^d(r^{i_1}W_1 + r^{i_2}W_2 + \left[h_{L^{i_1}}^{i_1}, h_{L^{i_2}}^{i_2}\right]W_3)\right) \tag{16}$$

Here, the combination is simply the linear weighted summation with parameter matrices $W_1, W_2, W_3$. $f^d$ is the decoder's output function and $\sigma$ is the sigmoid function. For set output, the loss function is multi-label loss defined as:

$$Loss_{set}\left(Y, \widehat{y}\right) = -\left(\sum_{y_l \in Y} \log \widehat{y_l} + \sum_{y_l \notin Y} \log\left(1 - \widehat{y_l}\right)\right) \tag{17}$$

For both settings, the decoder makes use of both memories' contents and encoders' final hidden values to produce the output. While memory contents represent the long-term knowledge, the encoder's hidden values represent the short-term information stored inside the controllers. Both are crucial to model late inter-view interactions and necessary for the decoder to predict the correct outputs.

### 3.4 Persistent Memory for Multiple Admissions

As mentioned earlier in Section 3.1, one unique property of healthcare is the long-term dependencies among admissions. Therefore, the output at the current admission $Y_a$ is dependent on the current and all previous admission's inputs $\left\{\left(X_{pa}^{i_1}, X_{pa}^{i_1}\right)\right\}_{pa=1}^{a}$. There are several ways to model this property. The simplest solution is to concatenate the current admission with previous ones to make up single sequence input for the model. This method causes data replication and preprocessing overhead. Another solution is to use recurrent neural network to model the dependencies. As in [3, 15], the authors use GRU and LSTM where each time step is fed with an admission. The admission is treated as a set of medical events and represented by a feature vector.

In our memory-augmented architecture, we can model this dependencies by using the memories to store information from previous admissions. In the original DNC, the memory content is flushed every time new data sample (i.e. new admission) is fed – this certainly loses the information of admission history. We modify this mechanism by keeping the memories persistent during a patient's admissions processing. That is, the content of memories is built up and modified during the whole history of a patient's admissions. The memories are only cleared prior to reading a new patient's record.

Persistent memories in our architecture play two important roles. First, because the number of events across admissions are large while memory sizes are moderate, the memory modules learn to compress efficiently the input views, keeping only essential information. This makes memory look-ups in the decoding process only limited to a fixed size of chosen knowledge. This is more compact and focused than attention mechanisms, in which the decoder has to attend to all events in the input. Second, each memory slot can store information of any event in the input views, which enables skip-connection reference in the decoding process, i.e., the decoder can jump to any input event, even the one in the farthest admission, to look for relevant information. The whole process of training our dual memory neural computer for healthcare data is summarized in Algorithm 1.

## 4 Results

In this section, we demonstrate the effectiveness of our proposed model DMNC on synthetic and real-world tasks. We use $DMNC_l$ and $DMNC_e$ to denote the late-fusion and early-fusion mode of our model, respectively. The data for real-world problems are real EMR data sets, some are public accessible. We make the source code of DMNC publicly available at https://github.com/thaihungle/DMNC.

### 4.1 Synthetic Task: Sum of Two Sequences

We conduct this synthetic experiment to verify our model performance and behavior. In this problem, the input views are two randomly generated sequence of numbers: $\{x_1^1, ..., x_L^1\}$, $\{x_1^2, ..., x_L^2\}$. Each sequence has $L$ integer numbers. $L$ is randomly chosen from range $[1, L_{max}]$ and the numbers are randomly chosen from range $[1, 50]$. The output view is also a sequence of integer numbers defined as $\{y_i = x_i^1 + x_{L+1-i}^2\}_{i=1}^L$, in which $y_i \in [2, 100]$. Note that this summation form is unknown to the model. During training, only the outputs are given.

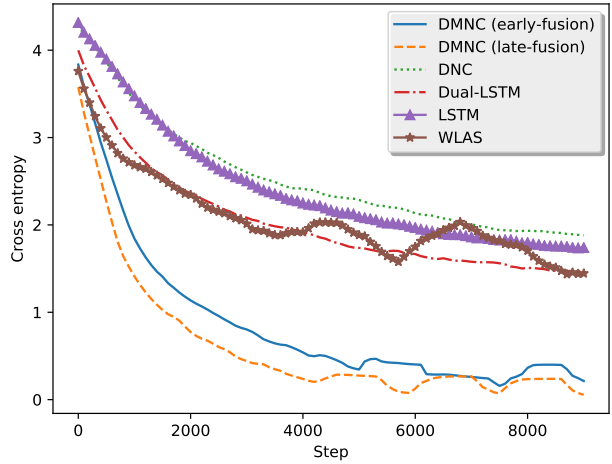

Figure 2: Training loss of sum of two sequences task. The training error curves have similar patterns.

Table 1: Sum of two sequences task test results. Max train sequence length is 10.

| Model | Accuracy (%) | | |
|---|---|---|---|
| | $L_{max} = 10$ | $L_{max} = 15$ | $L_{max} = 20$ |
| LSTM | 35.17 | 24.12 | 18.64 |
| DNC | 37.8 | 20.43 | 14.67 |
| Dual LSTM | 52.41 | 42.57 | 30.47 |
| WLAS | 55.98 | 43.29 | 32.49 |
| $DMNC_l$ | **99.76** | **98.53** | **78.17** |
| $DMNC_e$ | 98.84 | 93.00 | 69.93 |

Table 2: MIMIC-III data statistics.

| # of admissions | 42,586 | # of diag | 6,461 |
|---|---|---|---|
| # of patients | 34,594 | # of proc | 1,881 |
| Avg. view len | 53.86 | # of drug | 300 |

Table 3: Mimic-III drug prescription test results.

| Model | AUC | F1 | P@1 | P@2 | P@5 |
|---|---|---|---|---|---|
| Diagnosis Only | | | | | |
| Binary Relevance | 82.6 | 69.1 | 79.9 | 77.1 | 70.3 |
| Classifier Chains | 66.8 | 63.8 | 68.3 | 66.8 | 61.1 |
| LSTM | 84.9 | 70.9 | 90.8 | 86.7 | 79.1 |
| DNC | 85.4 | 71.4 | 90.0 | 86.7 | 79.8 |
| Procedure Only | | | | | |
| Binary Relevance | 81.8 | 69.4 | 82.6 | 80.1 | 73.6 |
| Classifier Chains | 63.4 | 61.7 | 83.7 | 80.3 | 71.9 |
| LSTM | 83.9 | 70.8 | 88.1 | 86.0 | 78.4 |
| DNC | 83.2 | 70.4 | 88.4 | 85.8 | 78.7 |
| Diagnosis and procedure | | | | | |
| Binary Relevance | 84.1 | 70.3 | 81.0 | 78.2 | 72.3 |
| Classifier Chains | 64.6 | 63.0 | 84.6 | 81.5 | 74.2 |
| LSTM | 85.8 | 72.1 | 91.6 | 86.8 | 80.5 |
| DNC | 86.4 | 72.4 | 90.9 | 87.4 | 80.6 |
| Dual LSTM | 85.4 | 71.4 | 90.6 | 87.1 | 80.5 |
| WLAS | 86.6 | 72.5 | 91.9 | 88.1 | 80.9 |
| $DMNC_l$ | 87.4 | 73.2 | **92.4** | 88.9 | **82.6** |
| $DMNC_e$ | **87.6** | **73.4** | 92.1 | **89.9** | 82.5 |

Because the output's number is the sum of two numbers from the two input views, we name the task as sum of two sequences. It should be noted that two input numbers in the summation do not share the same time step; hence, the problem is asynchronous. To learn and solve the task, a model has to read all the numbers from the two input sequences and discover the correct pair that will be used to produce the summation. Synchronous multi-view models certainly fail this task because they assume the inputs to be aligned. In the training phase, we choose $L_{max} = 10$, training for 10,000 iterations with mini batch size = 50. In the testing phase, we evaluate on 2500 random samples with $L_{max} = 10$, $L_{max} = 15$, $L_{max} = 20$ to verify the generalization of the models beyond the range where they are trained.

**Evaluations**: the baselines for this synthetic task are chosen as follows:

- View-concatenated sequential models: This concatenates events in input

Table 4: Example Recommended Medications by DMNCs on MIMIC-III dataset. Bold denotes matching against ground-truth.

| | |
|---|---|
| Diagnoses | Calculus Of Gallbladder (57411),Vascular disorders of male genital organs (60883), Abdominal Pain (78901), Poisoning By Other Tranquilizers (9695), Acute Myocardial Infarction Of Other Inferior Wall (41042), Hematoma Complicating (99812), Malignant hypertensive heart disease 40200), Dizziness and giddiness (7804), Venous (Peripheral) Insufficiency, Unspecified (45981), Hemorrhage Of Gastrointestinal Tract (5789) |
| Procedures | Coronary Bypass Of Three Coronary Arteries (3613), Single Internal Mammary Artery Bypass (3615), Extracorporeal circulation auxiliary to open heart surgery (3961), Insertion Of Intercostal Catheter For Drainage (3404), Operations on cornea(114) |
| Top 5 Ground-truth drugs (manually picked by experts) | Docusate Sodium (DOCU100L), Acetylsalicylic Acid (ASA81), Heparin (HEPA5I), Acetaminophen (ACET325), Potassium Chloride (KCLBASE2) |
| Top 5 Late-fusion Recommendations | **Docusate Sodium (DOCU100L)**, Neostigmine (NEOSI), **Acetaminophen (ACET325)**, Propofol (PROP100IG), **Potassium Chloride (KCLBASE2)** |
| Top 5 Early-fusion Recommendations | **Docusate Sodium (DOCU100L), Acetaminophen (ACET325), Potassium Chloride (KCLBASE2)**, Dextrose (DEX50SY), **Acetylsalicylic Acid (ASA81)** |

views to form one long sequence. This technique transforms the two-view sequential problem to normal sequence-to-sequence problem. We pick LSTM and DNC as two representative methods for this approach.

- Attention model WLAS [5]: This has a LSTM encoder per view, and attention is used for decoding, similar to that in machine translation [1]. The model is applied successfully in the problem of video sentiment analysis. To make it suitable for our tasks, we replace the encoders' feature-extraction layers in the original WLAS by an embedding layer. We choose this model as baseline since its architecture is somehow similar to ours. The difference is that we make use of external memories instead of attention mechanism.

- Dual LSTM: This model is the WLAS model without attention, that is, only the final states of encoders are passed into the decoder.

**Implementations:** For all models, embedding and hidden dimensions are 64 and 128, respectively. Word size for memory-based methods are 64. Memory size for the view-concatenated DNC and DMNC are 32 and 16, respectively. We double the memory size for view-concatenated DNC to account for the fact that the length of the input sequence is nearly double due to view concatenation. We use Adam optimizer with default parameters and apply gradient clipping size = 10 to train all models. Since output is a sequence, we use the cross-entropy loss function in Eq.(15). The evaluation metric used in this task is accuracy – the number of correct predictions over the length of output sequence.

**Results:** The training loss curves of the models are plotted in Fig. 2. The test average accuracy is summarized in Table 1. As clearly shown, overall the proposed model outperforms other methods by a huge margin of about 45%. Although dual LSTM and WLAS perform better than view-concatenated methods, it's too hard for non-memory methods to "remember" correctly pairs of inputs for later output summation. View-concatenated DNC even with double memory size still fails to learn the sum rule because storing two views' data in a single memory seems to mess up the information, making this model perform worst. Between two versions of DMNC, late-fusion mode is better perhaps due to the independence between two inputs' number sequences. This is the occasion where trying to model early cross-interactions damages the performance. The slight drop in performance when testing with $L_{max} = 15$ shows that our model really learns the sum rule. When $L_{max} = 20$, the input length is longer than the memory size, so even when DMNCs can learn the sum rule, they cannot store all input pairs for later summation. However, our methods still manage to perform better than any other baseline.

## 4.2 Drug Prescription Task

The data set used for this task is MIMIC-III, which is a publicly available dataset consisting of more than 52k EMR admissions from more than 46k patients. In this task, we keep all the diagnosis and procedure codes and only preprocess the drug code since the raw drug view's average length can reach hundreds of codes in an admission, which is too long given the amount of data. Therefore, only top 300 frequently used of total 4781 drug types are kept (covering more than 70% of the raw data). The final statistics of the preprocessed data is summarized in Table 2.

**Evaluations:** We compare our model with the following baselines:

- Bag of words and traditional classifiers: In this approach, each input view is considered as a set of events. The vector represents the view is the sum of one-hot vectors representing the events. These view vectors are then concatenated and passed into traditional classifiers: SVM, Logistic Regression, Random Forest. To help traditional methods handle multi-label output, we apply two popular techniques: Binary Relevance [10] and Classifier Chains [22]. We will only report the best model for each of the two techniques, which are Logistic Regression and Random Forest, respectively.

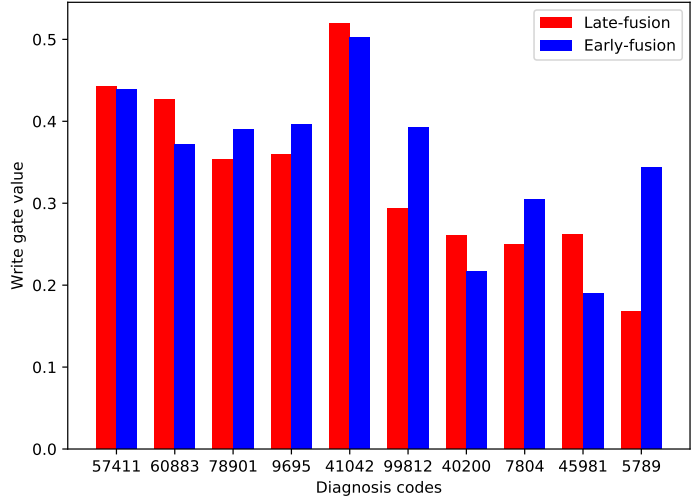

Figure 3: $M_1$'s $g_t^w$ over diagnoses. Diagnosis codes of a MIMIC-III patient is listed along the x-axis (ordered by priority) with the y-axis indicating how much the write gate allows a diagnosis to be written to the memory $M_1$.

- View-concatenated sequential models (LSTM, DNC), Dual LSTM and WLAS [5]: similar to those described in the synthetic task.

- Single-view models: To see the performance gains when making use of two input views, we also report results when only using one view for Binary Relevance, Classifier Chains, LSTM and DNC.

**Implementations:** We randomly divide the dataset into the training, validation and testing set in a $2/3 : 1/6 : 1/6$ ratio. For traditional methods, we use grid-searching over typical ranges of hyper-parameters to search for best hyper-parameter values. Deep learning models' best embedding and hidden dimensions are 64 and 64, respectively. Optimal word and memory size for DMNC are 64 and 16, respectively. The view-concatenated DNC shares the same setting except the memory size is doubled to 32 memory slots. Since the output in this task is a set, we use the multi-label loss function in Eq.(17) for deep learning methods. We measure the relative quality of model performances by using common multi-label metrics, Area Under the ROC Curve (AUC) and F1 scores, both of which are macro-averaged. Similar results can be achieved when using micro-averaged so we did not report them here. In practice, precision at $k$ (P@$k$) are often used to judge the treatment recommendation quality. Therefore, we also include them ($k = 1, 2, 5$) in the evaluation metrics.

**Results:** Table 3 shows the performance of experimental models on aforementioned performance metrics. We can see the benefit of using two input views instead of one, which helps improve the model performances. Traditional methods clearly underperform deep learning methods perhaps because these methods are hard to scale when there are many output labels and the inputs in our problem are not bag-of-words. Among deep learning models, our proposed ones consistently outperform others in all type of measurements. Our methods demonstrate 1-2% improvements over the second runner-up baseline WLAS. The late-fusion mode seems suitable for certain type of metrics, but overall, the early-fusion mode is the winner, highlighting the importance of modeling early inter-view interactions.

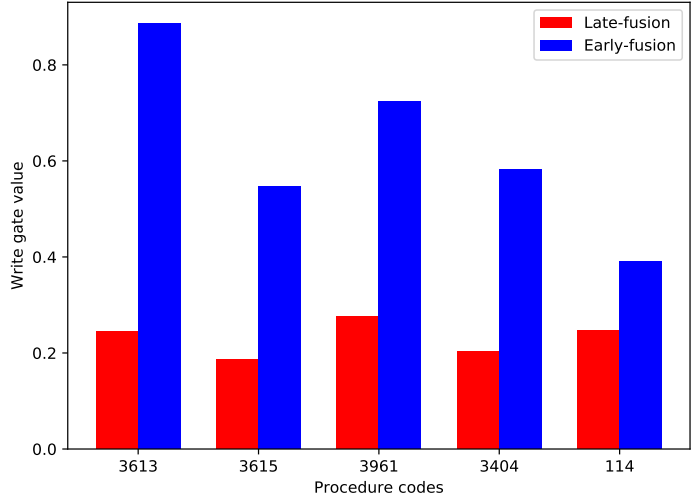

Figure 4: $M_2$'s $g_t^w$ over procedures. Medical procedure codes of a MIMIC-III patient is listed along the x-axis (in the order of executions) with the y-axis indicating how much the write gate allows a procedure to be written to the memory $M_2$.

**Case study:** In Table 4, we show an example of drugs prescribed for a patient given his current diagnoses and procedures. The patient had serious problems with his bowel as described in the first four diagnoses. The next three diagnoses are also severe relating to his heart problems while the remaining diagnoses are less urgent. It seems that heart-related diagnoses later led to heart surgeries listed in the procedure codes. Both modes of DMNC predict correctly the drug Docusate Sodium used to cure urgent bowel symptoms. Relating to heart diseases and surgeries, our models predict closely to expert's choices. Potassium Chloride is necessary for a healthy heart. Acetaminophen and Propofol are commonly used during surgeries. However, some heart medicine such as Heparin is missed by the two models. Figs. 3 and 4 demonstrate the "focus" of the two memories on diagnosis and procedure view, respectively. The higher the write gate values, the more information of the medical codes will be written into the memories. We can see both modes pay less attention on last diagnoses corresponding to less severe symptoms. Compared to the late-fusion, the early-fusion mode keeps more information on procedures, especially the heart-related events. This may help increase the weight on heart-related medicines and enable it to include Acetylsalicylic Acid, a common drug used after heart attack in the top recommendations.

Table 5: Regional hospital test results. P@K is precision at top K predictions in %.

| Model | Diabetes | | | Mental | | |
|---|---|---|---|---|---|---|
| | P@1 | P@2 | P@3 | P@1 | P@2 | P@3 |
| DeepCare | 66.2 | 59.6 | 53.7 | 52.7 | 46.9 | 40.2 |
| WLAS | 65.9 | 60.8 | 56.5 | 51.8 | 48.9 | 45.7 |
| $DMNC_l$ | 66.5 | **61.3** | **57.0** | 52.7 | 49.4 | 46.2 |
| $DMNC_e$ | **67.6** | 61.2 | 56.9 | **53.6** | **50.0** | **47.1** |

## 4.3 Disease Progression Task

Data used in this task are two chronic cohorts of diabetes and mental EMRs collected between 2002-2013 from a large regional hospital in Australia. Since we want to predict the next diagnoses for a patient given his or her history of admission, we preprocessed the datasets by removing patients with less than 2 admissions, which ends up with 53,208 and 52,049 admissions for the two cohorts. In this data set, procedures and medicines are grouped into intervention codes, together with diagnosis codes forming a patient's admission record. The number of diagnosis and intervention codes are 249 and 1071, respectively. We follow the same preprocessing steps and data split as in [15]. Different from MIMIC-III, a patient record suffering from chronic conditions often consists of multiple admissions, which is suitable for the task of predicting disease progression. The average number and the maximum number of admission per patient are 5.35 and 253, respectively.

**Evaluations:** For comparison, we choose the second best-runner in our previous experiments WLAS and the current state-of-the-art DeepCare [15] as the two baselines.

**Implementations:** We use the validation data set to tune the hyper-parameters of our implementing methods and have the best embedding and hidden dimensions are 20 and 64, respectively. The word and memory size for DMNC are found to be 32 and 32, respectively. For performance measurements, we use P@$k$ metric ($k = 1, 2, 3$) to make it comparable with DeepCare's results reported in [15].

**Results:** We report the results on test data of models for disease progression task in Table 5. For both cohorts, our proposed model consistently outperforms other methods and the performance gains become larger as the number of predictions increase. Compared to DeepCare which uses pre-trained embeddings and time-intervals as extra information, our methods only use raw medical codes and perform better. This emphasizes the importance of modeling view interactions at event level. The late-fusion DMNC seems to perform slightly better than the early-fusion DMNC in the diabetes cohort, yet overall, the latter is the better one, which again validates its ability to model all types of view interactions.

## 5 Conclusions

This paper proposes a novel deep learning architecture for asynchronous two-view sequential learning called Dual Memory Neural Computer (DMNC). Under our design, each input view is assigned a neural controller to encode and store its events to a dedicated memory. After all input views are stored, a decoder will access the memories and synthesize the read contents to produce the final output. Moreover, our model is equipped with two modes of memories, enabling it to model comprehensively all types of view interactions. Through extensive experiments, DMNC is compared with various baselines and consistently shows better performance on three tasks: sum of two sequences, drug prescription and disease progression. Future works will focus on generalizing our model to multi-input multi-output settings and extending the range of applications to bigger problems such as multi-media and multi-agent systems.